\documentclass[runningheads]{llncs}

\usepackage{eccv}

\usepackage{eccvabbrv}

\usepackage{graphicx}
\usepackage{booktabs}
\usepackage{utfsym}
\usepackage{multirow}
\usepackage{booktabs}
\usepackage{makecell}
\usepackage{ulem}
\usepackage{wrapfig}
\usepackage{bbding}
\renewcommand\footnotemark{}

\usepackage[accsupp]{axessibility}  

\usepackage{hyperref}

\usepackage{orcidlink}

\begin{document}


\title{Taming Lookup Tables for Efficient Image Retouching} 


\author{Sidi Yang\thanks{* ~ Contributed equally}\inst{1}\textsuperscript{*} \and
Binxiao Huang\inst{2}\textsuperscript{*}    \and
Mingdeng Cao\inst{3} \and
Yatai Ji\inst{1} \and
Hanzhong Guo\inst{4} \\
Ngai Wong\inst{2} \and
Yujiu Yang\inst{1}\thanks{\textsuperscript{\Envelope} ~ Corresponding author}\Envelope
}

\authorrunning{S.~Yang, et al.}

\institute{Tsinghua University~ \email{\{yangsd21,jyt21\}@mails.tsinghua.edu.cn} \quad \email{yang.yujiu@sz.tsinghua.edu.cn}\and 
The University of Hong Kong~ \email{\{bxhuang,nwong\}@eee.hku.hk}\and 
The University of Tokyo~ \email{mingdengcao@gmail.com} \and
Renmin University of China~ \email{guohanzhong@ruc.edu.cn}}

\maketitle

\begin{abstract}
The widespread use of high-definition screens in edge devices, such as end-user cameras, smartphones, and televisions, is spurring a significant demand for image enhancement. Existing enhancement models often optimize for high performance while falling short of reducing hardware inference time and power consumption, especially on edge devices with constrained computing and storage resources. To this end, we propose \textbf{I}mage \textbf{C}olor \textbf{E}nhancement \textbf{L}ook\textbf{U}p \textbf{T}able (\textbf{ICELUT}) that adopts LUTs for extremely efficient edge inference, without any convolutional neural network (CNN). During training, we leverage pointwise ($1\times 1$) convolution to extract color information, alongside a split fully connected layer to incorporate global information. Both components are then seamlessly converted into LUTs for hardware-agnostic deployment. ICELUT achieves near-state-of-the-art performance and remarkably low power consumption. We observe that the pointwise network structure exhibits robust scalability, upkeeping the performance even with a heavily downsampled $32\times32$ input image. These enable ICELUT, the \textit{first-ever} purely LUT-based image enhancer, to reach an unprecedented speed of 0.4ms on GPU and 7ms on CPU, at least one order faster than any CNN solution. Codes are available at \href{https://github.com/Stephen0808/ICELUT}{https://github.com/Stephen0808/ICELUT}.
\end{abstract}

\section{Introduction}
\label{sec:intro}

In digital camera imaging, adverse shooting conditions and limited computing power can cause a decline in image quality. To meet aesthetic preferences, the conventional enhancement process involves using expert-designed cascade modules for exposure compensation, saturation adjustment, and tone mapping. However, these laborious and inflexible adjustments often result in unsatisfactory images. To overcome this, deep learning methods have gained popularity in automatically retouching images.

\begin{figure}[h] 
    \begin{center}
        \includegraphics[width=0.7\textwidth]{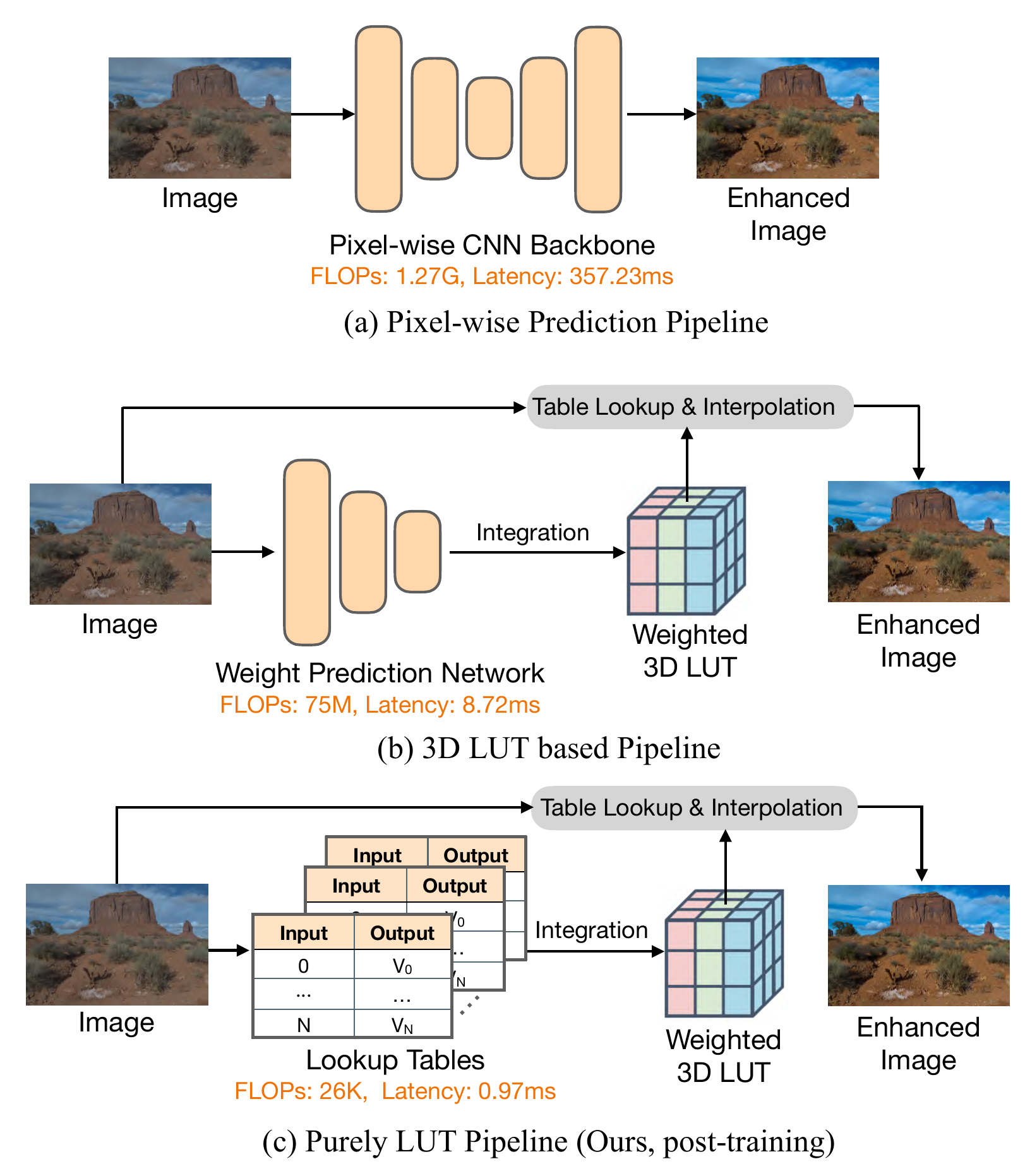}
        \caption{Three image enhancement pipelines. FLOPs and latency are measured on the CPU for the (orangish) backbones. CSRNet\cite{he2020conditional} and CLUT\cite{zhang2022clut} are chosen as representatives for the \textbf{(a)} end-to-end and \textbf{(b)} 3D LUT methods, and our approach, developed post-training, is \textbf{(c)} purely LUT-based.}
        \label{fig:intro}
    \end{center}
\end{figure}

Previous works can be categorized into two classes: 1) image-to-image network, which directly transforms the input image to its enhanced version, and 2) predicting a 3D lookup table (LUT) to map input pixels (i.e., RGB values) to their enhanced counterparts. In the first class, convolution kernels are applied to process the overall image pixels. In this pipeline, since all pixels within the kernel are involved, the computational burden is high. 
To overcome this hurdle, \cite{zeng2020learning} proposed a 3D LUT-based method that converts the pixel prediction network into a weight prediction network, which is trained to predict a weight tensor for weighting a series of basis 3D LUTs. Since the weights correspond to the brightness, color, and tones of images, the network only needs to process a downsampled image with significantly reduced computation. This makes an important step toward real-time image inference. However, most edge or portable devices have limited compute or power budgets, making it demanding to perform resource-intensive computations for each image inference. As shown in Fig.~\ref{fig:intro}, both the pixel prediction and weight prediction networks, due to their convolutional neural network (CNN) nature, exhibit high Floating Point Operations (FLOPs). In contrast, a LUT approach constitutes a cost-effective and hardware-friendly data structure that requires only a position index to retrieve the output directly. Nonetheless, while LUTs are efficient for inference, a tradeoff exists between their representation power and storage requirement. Specifically, a larger feature vector (used as address indices) enhances representation, but this incurs an exponential growth in LUT size, which can often be prohibitive. A natural question arises: \textit{Can we transfer the neural network to a reasonably sized LUT for saving FLOPs and reducing latency without compromising image quality?} Luckily, this work, \textbf{\textit{for the first time}}, provides an affirmative answer to a pure LUT-based image enhancer.

Since the LUT size scales exponentially with the input dimension~\cite{jo2021practical,li2022mulut,ma2022learning},
we first studied the impact of receptive field size (spatial) and the number of input channels (depth) on performance. To this end, we found that the model can achieve high performance even with a tiny receptive field, but the absence of the three RGB channels severely hampers performance. Therefore, \textbf{during training}, we propose a network with fully pointwise convolution layers for feature extraction, where the receptive field size stays at $1\times1$. 
Each input pixel with three channels can be uniquely mapped to an output through a LUT. To efficiently process 8-bit (INT8) color images, we employ two parallel branches to separately process the 4 most significant bits (MSBs), denoted $I_{MSB}$, and
the 4 least significant bits (LSBs), denoted $I_{LSB}$. It reduces the LUT addresses (viz. input indices) from $256^3$ to $2 \times 16^3$.
Furthermore, the feature output from the fully pointwise network is pooled and fed into a fully connected (FC) layer for integrating the global information. However, the typical FC input feature dimension is inevitably large to be the address indices of a practically sized LUT. To overcome this curse of dimensionality, we propose a split FC layer that divides its input features into groups of small length and uses separate FC layers for each group. It reduces the memory from $(V)^{C}$ to $L\times(V)^{K}$, where $V$, $C$, $L$ and $K$ stand for the possible input values, channel numbers, group length and number of groups, respectively, with $C=L\times K$. All outputs are summed to get the weight vector, which is used for linearly combining the basis LUTs into a 3D LUT for final table lookup and interpolation. \textbf{After training}, we transfer such CNN+FC backbone into LUTs, leading to merely table lookup operations with minimal FLOPs during inference.

Input resolution is crucial for latency and FLOPs in image processing. Traditional 3D LUT methods downsample images to $256\times256$ for faster real-time inference. Generally, smaller resolutions result in lower latency and FLOPs. Our proposed architecture explores input resolution and receptive field, finding that a pointwise receptive field minimizes performance loss even with significantly downsampled images. This enables our network to use $32\times32$ downsampled inputs while maintaining similar performance to the original resolution. Our paper's main contributions are threefold:
\begin{enumerate}
    \item We find the channel number is vital for image retouching, and designing a network with fully pointwise convolution kernels is favorable for LUT conversion.
    \item We reveal a small receptive field instills robustness to low-resolution input for training and inference. An unprecedented $32 \times32$ downsampled image is used in our network for extreme speed with minimal performance drop.
    \item Our purely LUT scheme achieves a remarkable 0.4ms (7ms) on GPU (CPU) with near-state-of-the-art performance and reduces the power consumption to a negligible level compared to CNN schemes.
\end{enumerate}

\section{Related works}
\label{sec:related}
\subsection{Learning-based image enhancement}
Since the introduction of the large-scale dataset MIT-Adobe FiveK~\cite{bychkovsky2011learning}, which contains input and expert-retouched image pairs, numerous learning-based enhancing algorithms have emerged to propel this field. Generally, learning-based methods can be divided into two categories: image-to-image translation and physics-based modeling.

The first category treats this task as an image-to-image translation, directly learning the end-to-end mapping between the input and its enhanced image without explicitly modeling intermediate parameters.~\cite{afifi2019color} and~\cite{kim2020pienet} both use UNet-like networks to predict the enhanced results, while~\cite{chen2018deep} utilizes two-way generative adversarial networks (GANs) trained with unpaired retouched data for the scarcity of expert-retouched data collection. To further complement feature extraction,~\cite{wang2019underexposed} explores the connection of illumination with this task and designs an encoder-decoder-based network for image enhancement.

In the second category, domain prior knowledge is utilized as the guided information for model design since it is often simpler to predict the transformation from input to output rather than predicting the output directly~\cite{shih2013data}. By viewing the enhancement procedure as a nonlinear transformation,~\cite{afifi2019color} predicts polynomial mapping functions while the one-dimensional RGB curves are approximated in~\cite{moran2021curl,kim2020global}. 
To further accelerate inference, HDRNet~\cite{gharbi2017deep} performs most of the inference on a low-resolution copy of the input and applies bilateral filters with neural networks, achieving millisecond processing for 1080p resolution images. However, the generally required extra parameters (e.g., $3\times3$ convolutions) are not specialized for image enhancement, leading to large implementation redundancy. To this point,~\cite{he2020conditional} uses a lightweight backbone for pixel processes and designs a block for global feature extraction and merging, which is orders of magnitude smaller than other methods. Recently, \cite{ouyang2023rsfnet} adapts region maps and human-understandable
filter arguments to achieve fine-grained enhancement. 

\subsection{3D LUT-based image enhancement}
The first 3D LUT-based method was proposed in~\cite{zeng2020learning}, which converts the pixel-level prediction into coefficients prediction. Benefiting from the regularized LUT, the lightweight network only predicts a set of coefficients for weighting the trainable LUTs. The 3D LUT is extended to other tasks and scenes along this line. By analyzing channel coherence, CLUT~\cite{zhang2022clut} applies a transformation matrix to compress the LUT adaptively. \cite{wang2021real} embeds the pixel-wise category information into the combination of multiple LUTs, while~\cite{yang2022seplut} separates a single color transform into a cascade of component-independent and component-correlated sub-transforms instantiated as 1D and 3D LUTs, respectively. Recently, \cite{liu20234d} utilizes 4D LUT to achieve content-dependent enhancement.

\subsection{Replacing CNN with LUT}
Although 3D LUT-based methods can achieve remarkable efficiency, the CNN coefficient prediction network still consumes large computational resources compared to table lookup. Recently, several works have delved into converting neural networks into LUTs to bypass computation.~\cite{jo2021practical} first converts the CNN with a limited receptive field into LUTs for super-resolution. In this paradigm, the network receives restricted pixels (e.g., $2\times2$) and predicts corresponding super-resolution pixels during training. The CNN is then transferred into a LUT by traversing all possible combinations of input values in the receptive field. During inference, the high-resolution pixels are retrieved by the input pixel values serving as address indices to the LUT. The subsequent work~\cite{ma2022learning,li2022mulut,huang2023hundred} adopts serial LUTs for enlarging the receptive field and achieves a large PSNR improvement. To fully exploit the spatial pixels while avoiding exponential blow-up of the LUT size, these methods only build independent LUTs for a single input channel, i.e., treating the RGB channels unanimously. However, different from super-resolution, which focuses on restoring high-frequency details utilizing spatial information, image color enhancement requires the preservation of color information due to channel interaction. Hence, the aforementioned channel-agnostic LUTs are not viable for image enhancement.


\section{Method}
\subsection{3D LUT preliminaries}
3D LUT is a highly efficient tool for real-time image enhancement, which models a nonlinear 3D color transform by sparsely sampling it into a discretized 3D lattice. Previous methods have tried designing models for learning image-adaptive 3D LUTs, utilizing a lightweight CNN backbone to predict weights for fusing a series of basis 3D LUTs to form an image-dependent 3D lattice $V = \{(V_{r, (i, j, k)}, V_{g, (i, j, k)}, V_{b, (i, j, k)})\}_{i, j, k =0,1,...,M-1}$, where $M$ is the number of bins in each color channel. Each element $V_{(i,j,k)}$ defines an indexing RGB color $\{r^I_{(i, j, k)}, g^I_{(i, j, k)}, b^I_{(i, j, k)}\}$ and the corresponding transformed output RGB color $\{r^O_{(i, j, k)}, g^O_{(i, j, k)}, b^O_{(i, j, k)}\}$. Once a 3D lattice is sampled, an input pixel looks up its nearest index points according to its color and computes its mapping output via, say, trilinear interpolation.


However, the heavy computation in the CNN still remains a burden for inference on edge devices. Here we use the extremely lightweight method, CLUT~\cite{zhang2022clut}, as an example. We observe $35\times$ latency and $15\times$ FLOPs in the CNN forward pass versus the subsequent 3D LUT mapping (viz. interpolation and table lookup) as shown in Table~\ref{tab:latency}, which echoes that table lookup is substantially more efficient. A question follows: Is it possible to convert a CNN into a purely LUT-based model that possesses \textbf{\textit{fast inference speed}}, \textbf{\textit{cost effectiveness}} and \textbf{\textit{high performance}}? Ideally, the enhanced images could be generated only by table lookup without any neural network computation.

\begin{table}[th]
    \centering
    \caption{Latency analysis of 3D LUT-based method. Here we use the CLUT~\cite{zhang2022clut} inference block as an example. Results are tested on 480p images on an NVIDIA GeForce 3090 card.}
    \begin{tabular}
    {p{36mm}<{\centering} p{20mm}<{\centering} p{20mm}<{\centering}}
        \toprule[1.1pt]
        Block & Latency (ms) &FLOPs (M)\\
        \hline
        Weight Prediction & 2.15 &230\\
        3D LUT mapping& 0.06 &15\\
        \bottomrule[1.1pt]
        
    \end{tabular}
    \label{tab:latency}
\end{table}

Converting the computation in a neural network to a LUT is nontrivial. On the one hand, a complete LUT has to store all possible combinations of input pixel values, while an exponential relationship is evident between the number of input pixels and the LUT size. It means the number of input pixels must be restricted. On the other hand, every input combination corresponds to a unique output, implying that the network's receptive field is equivalent in size to the input dimension of the LUT. Table~\ref{tab:tradeoff} shows the relationship between LUT dimension and storage.

\begin{table}[th]
    \centering
    \caption{LUT size versus receptive field (RF) and channel numbers.}
    \begin{tabular}
    {p{20mm}<{\centering} p{15mm}<{\centering} p{20mm}<{\centering} p{30mm}<{\centering}}
        \toprule[1.1pt]
        RF& \#Channels &LUT Dim.& LUT size\\
        \hline
        $1\times1$ & 1 & 1D &256 B\\
        $1\times1$& 3& 3D &16 MB\\
        $2\times2$ & 1 & 4D &4 GB\\
        $2\times2$ & 3 & 12D & $7.21\times 10^{16}$ TB\\
        $k\times k$ & c & $k \times k \times c$D & $(2^{8})^{k \times k \times C}$ B\\
        \bottomrule[1.1pt]
    \end{tabular}
    \label{tab:tradeoff}
\end{table}

\begin{table}[th]
    \centering
    \caption{Ablating CNNs on FiveK datasets. With an enlarged RF, ConvNet-1C-$2\times2$ achieves a limited improvement in PSNR over ConvNet-1C-$1\times1$. While with the three channels involved in the input, ConvNet-3C-$1\times1$ largely outperforms the performance of ConvNet-1C-$1\times1$.}
    \resizebox{0.9\columnwidth}{!}{
    \begin{tabular}
    {p{30mm}<{\centering} p{30mm}<{\centering} p{14mm}<{\centering} p{18mm}<{\centering}}
    \toprule[1.1pt]
      Method & \#Channels $\times$ (RF) & LUT Dim.&PSNR (dB) \\
      
      \hline
        ConvNet-1C-$1\times1$ & $1\times (1\times1)$ & 1D & 24.16 \\
        ConvNet-1C-$2\times2$ & $1\times (2\times2)$ & 4D & 24.25 \\
        ConvNet-3C-$1\times1$ & $3\times (1\times1)$ & 3D &25.10 \\
        ConvNet-3C-$2\times2$ & $3\times(2\times2)$ & 12D & 25.16 \\
      \bottomrule[1.1pt]
    \end{tabular}
    }
    \label{tab:RF}
\end{table}

\begin{figure}[t]
    \centering
    \includegraphics[width=1.0\columnwidth]{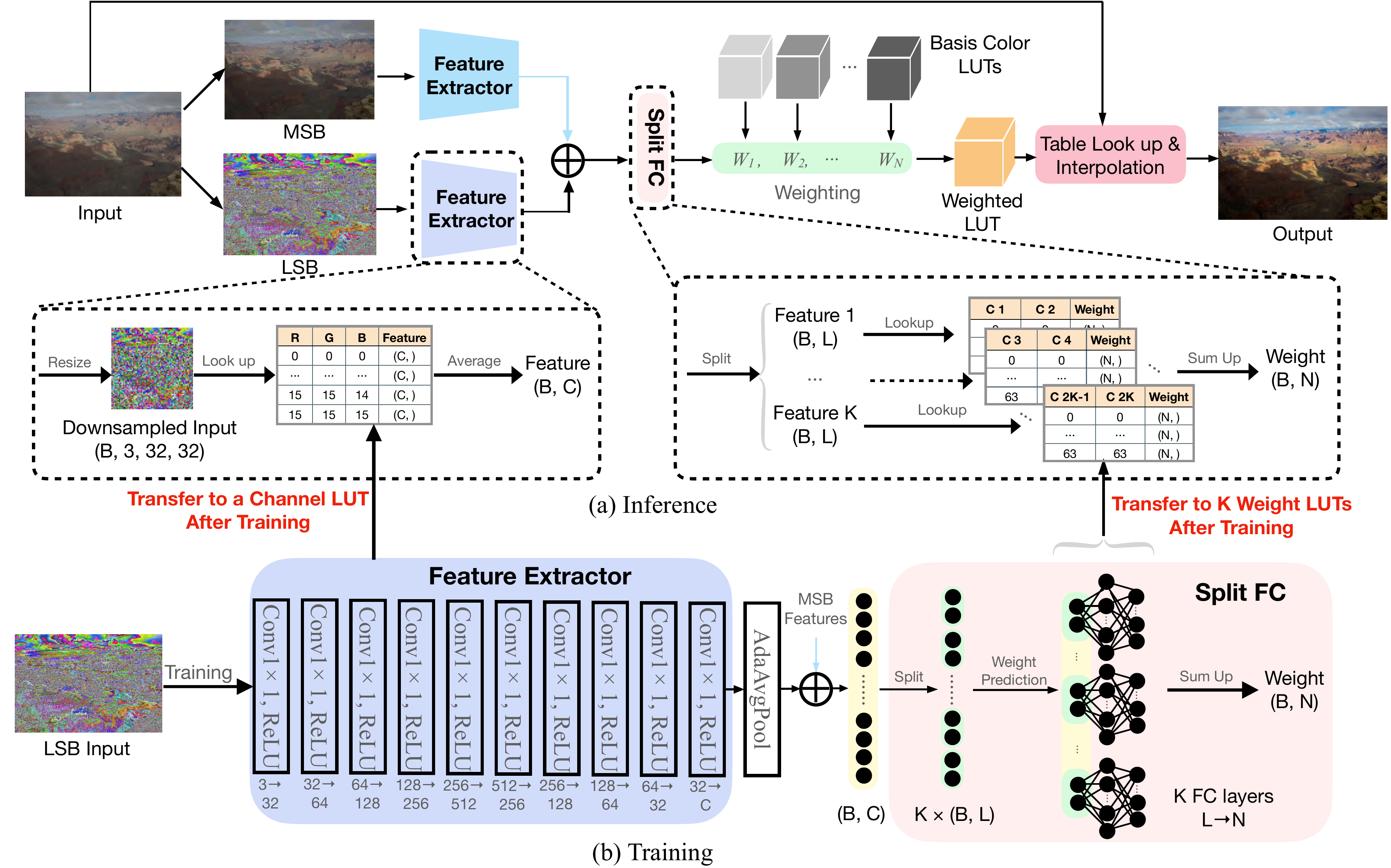}
    \caption{Overall architecture of the proposed ICELUT. Our model first employs a two-branch structure to parallelly process the MSB and LSB maps and then uses a fully pointwise network with a restricted receptive field to extract features. Furthermore, a split FC layer is utilized to fuse the global information for predicting the weights to combine the 3D LUTs for table lookup and interpolation. Note that the feature extractor and split FC, once trained, are transferred to lookup tables for purely LUT inference.}
    \label{fig:IELUT}
\end{figure}

To exploit the potential of the LUT, we conduct a preliminary ablation experiment to evaluate the importance of the receptive field and channel depth. Referring to Table~\ref{tab:RF}, we observe that the expansion of channels results in a significant improvement in PSNR compared to the increase in spatial dimensions. This reveals that the interdependent information of channels is vital for image enhancement, whereas the contribution of nearby pixels is limited. For example, smooth regions such as the sky or beach often possess similar pixels with spatial consistency but different channel information. In light of this, we design an ultra-lightweight network with two key attributes: 1) the ability to capture essential image information for producing high-quality enhanced images and 2) efficiency for conversion into LUTs, enabling accelerated inference.

\subsection{Training network}
The proposed \textbf{I}mage \textbf{C}olor \textbf{E}nhancement LUT (\textbf{ICELUT}) network is shown in Fig.~\ref{fig:IELUT}. It consists of two CNNs and a split FC layer during training. The 8-bit input pixels are separated into two maps, $I_{MSB}$ with 4 MSBs and $I_{LSB}$ with 4 LSBs, and fed into the two parallel CNN branches.

\begin{figure*}[htp]
    \centering
    \includegraphics[width=0.98\textwidth]{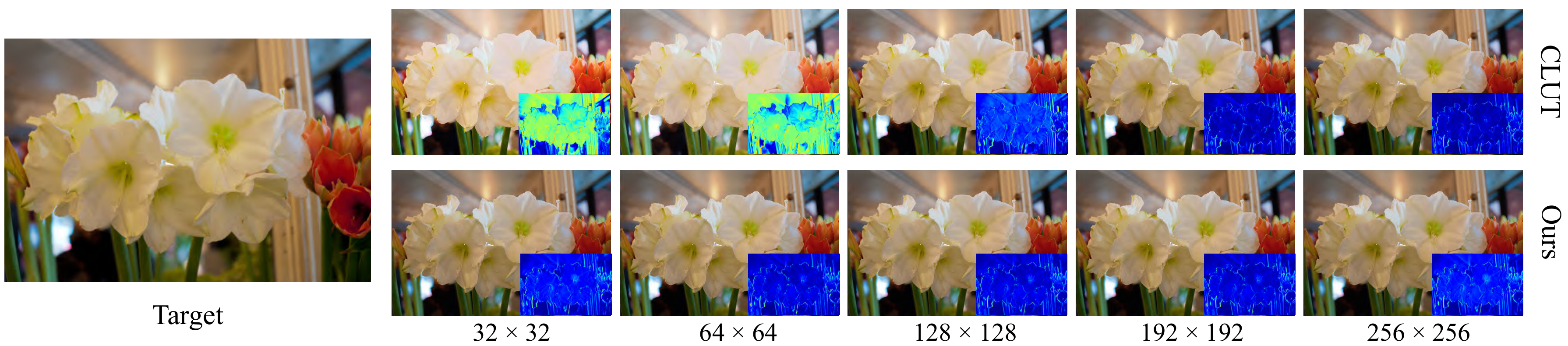}
    \caption{Visualization of results at different inference scales. The bottom right shows the error map with the target image. Brighter areas indicate larger absolute errors.}
    \label{fig:vis_scale}
\end{figure*}

\subsubsection{CNN backbone}
Since the number of input pixels decides the LUT size and channel depth mainly affects the network performance, we constrain the convolution kernel shape to a very small spatial size and full channel depth. We adopt six $1\times1$ convolution layers followed by ReLU activation. The depth of convolution kernels is set to 3 for processing all of the RGB channels. Since the receptive field (RF) of the repeated $1\times1$ convolution layers is still $1\times1$, we use an adaptive average pooling layer for aggregating the spatial features and compressing the feature map into $1\times1$. This pooling module plays an important role here in fusing the global information, which complements the limited local information extracted by former $1\times1$ convolution layers. 

\subsubsection{Split fully connected layer}\label{sec:sfc}
Simply predicting weights by pooling the feature map leads to sub-optimal results. The adaptive average pooling, which is a non-learning module, fuses global information into the feature map by coarsely compressing the feature map. For an abundant representation, we use an FC layer to map the feature into weights. The input of the FC layer corresponds to the number of output channels $C$ of the CNN backbone. Generally, more output channels represent a more abundant representation. However, the LUT size is exponential to channel number $C$: 
\begin{equation}
    S = V^C\times N,
\end{equation}
where $S$ is the size of LUT, $V$ denotes the possible values in a dimension, and $N$ is the output dimension. When $C>4$, $V=64$ and $N=20$, the LUT is generally beyond 1GB. To avoid substantial memory consumption, we design the split fully connected (SFC) layer. First, we split the input tensor into $K$ groups, and each group contains $L$ values from channel features. Then, we apply a vanilla FC layer to map each 2D feature into a weight of length $N$. These predicted weights are added at the end for weighting the basis LUTs to build a 3D weighted LUT. When we set $L=2$ and $K=C/2$, this manipulation reduces the memory from $(V)^C\times N$ to $(C/2)\times (V)^2 \times N$. Subsequently, when $C=10$, $V=64$ and $N=20$, the LUT size is equal to 400KB, which is orders smaller than the vanilla FC layer. 

\subsection{Transferring to LUT}\label{sec:sfc2}
The aforementioned network design paves the way for transferring to LUT. For the CNN backbone, we build a 3D LUT for the three channels while the RF size is 1 due to the pointwise convolution layers. Since the size of Channel LUT is comparably small, we do not quantize the output anymore. For the SFC layer, we build $K$ 2D LUTs for the feature mapping, named Weight LUTs. The input values are used to index the LUT, with the corresponding output value stored at that address. Note that the formats of the stored output values in the Channel LUT and Weight LUT are FP32 and INT8, respectively. 

For indexing the values in Weight LUTs, we quantize the input continuous values (FP32) into discrete values for building the indices of Weight LUTs:
\begin{equation}\label{eq1}
    Q = \text{Clamp}\left(\frac{\lfloor U \times \Delta s \rfloor}{\Delta s}, -R, R-(\frac{1}{\Delta s})\right),
\end{equation}
where $Q$ denotes the quantized value, $\Delta s$ denotes the sampling interval, $R$ denotes the offset, and $U$ denotes the output (FP32) of the average pooling module. Clamp$(\cdot, min, max)$ is to clip values outside $min$ and $max$. We compute the output values of the learned split FC by traversing all possible input combinations and saving them into the Weight LUTs. During inference, we seamlessly convert the quantized values into indices (integers):
\begin{equation}\label{eq2}
    I = \lfloor (Q + R) \times \Delta s \rfloor,
\end{equation}
where $I$ represents the index of Weight LUTs. In the paper, we set $\Delta s =2$ and $R=16$, which has 64 values in one dimension of Weight LUTs. While quantization brings about a 0.05dB performance drop, it significantly reduced the scale of storage.

\subsection{Speeding up inference}

Our main goal is to predict a set of weights for weighting the basis LUTs to produce the final 3D LUT. The CNN weight predictor aims to understand the global context, such as brightness, color, and tones of the image, to output content-dependent weights. Therefore, it only needs to work on the downsampled input image to largely lessen the computational cost~\cite{zeng2020learning}. Given an input image of any resolution, previous works simply used bilinear interpolation to downsample it to $256\times256$ for high efficiency. Nevertheless, as the input size increases, the network's computational workload, or table lookup operation, also escalates. Hence, a reduction in the image resolution can directly speed up inference and provide cost savings. Subsequently, we compare the performance of ours and existing methods. All methods are trained under the $256\times256$ resolution and tested for various downsampled scales. The results of CLUT are shown in Fig.~\ref{fig:vis_scale}.

\begin{figure}[htp] 
    \begin{center}
        \includegraphics[width=0.65\textwidth]{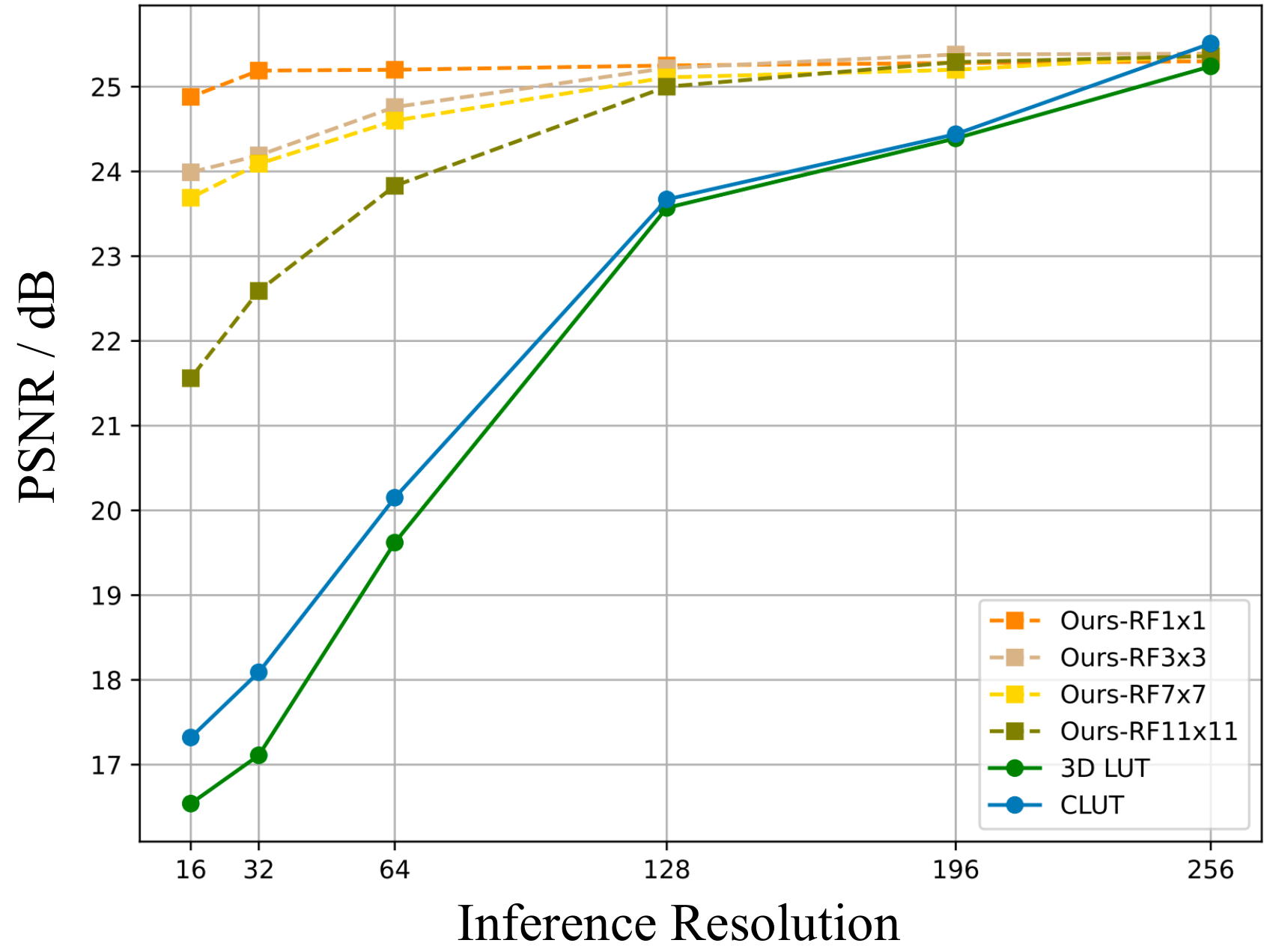}
        \caption{PSNR w.r.t. different resolutions on FiveK dataset.}
        \label{fig:res_comp}
    \end{center}
\end{figure}

We observe that at high resolutions, different models consistently performed well. However, as the resolution decreases, the models using a large RF (viz. 3D LUT and CLUT) suffer a sharp performance drop, while ICELUT (with a $1\times1$ RF) maintains a high level of performance.



        


To further verify the reason for the performance drop, we replace the first layer of our CNN backbone with $3\times 3$, $7\times7$, and $11\times 11$ convolution kernels to enlarge the RF with other settings being fixed. The results, shown in Fig.~\ref{fig:res_comp}, clearly indicate that networks with a larger RF suffer from a more severe performance drop. 



\section{Experiments}

\subsection{Datasets}
We evaluate the proposed ICELUT using two public datasets: MIT-Adobe FiveK~\cite{bychkovsky2011learning} and PPR10K~\cite{liang2021ppr10k}. The MIT-Adobe FiveK dataset is a well-known photo retouching dataset comprising 5,000 RAW images. We follow the common practice established in recent works~\cite{zeng2020learning,zhang2022clut}, by selecting the version retouched by expert C as the ground truth. We divide this dataset into 4,500 image pairs for training and 500 for testing. To expedite the training process, we downsized the images to 480p resolution, where the shorter side is resized to 480 pixels. The PPR10K dataset contains an extensive collection of 11,161 high-quality RAW portrait photos. In our experiments, we use all three retouched versions as the ground truth in three separate experiments. Adhering to the official dataset split as in~\cite{liang2021ppr10k}, we divide the dataset into 8,875 pairs for training and 2,286 for testing. For the sake of efficiency and due to limited disk space, we perform these experiments using the 360p version of the dataset. Note that the performance data in \cite{liang2021ppr10k} are not fair in the discussion of runtime and FLOPs since the 3D LUT trained in \cite{liang2021ppr10k} used a much larger backbone, ResNet18\cite{he2016deep} (about 11M). To this point, we have retrained 3D LUT using the original tiny backbone in \cite{zeng2020learning}.

\begin{table}[h] 
    \begin{center}
    \caption{Quantitative comparison (FiveK).}
    \begin{tabular}
    {p{30mm}<{\centering} p{20mm}<{\centering} p{20mm}<{\centering} p{18mm}<{\centering}}
        \toprule[1.1pt]
        Method &PSNR $\uparrow $& SSIM $\uparrow$ &$\Delta E \downarrow$ \\
        \hline
        UPE\cite{wang2019underexposed} & 21.88 & 0.853 & 10.80\\
        DPE\cite{chen2018deep} & 23.75 & 0.908 & 9.34\\
         HDRNet\cite{gharbi2017deep} & 24.32 &0.912&8.49\\
         CSRNet\cite{he2020conditional} & 25.21 &0.923 & 7.70\\
         3D LUT\cite{zeng2020learning} & 25.19 &0.912 & 7.61\\
         CLUT\cite{zhang2022clut} & 25.53 & 0.926 & 7.46\\
         ICELUT & 25.27 & 0.918 & 7.51\\
        \bottomrule[1.1pt]
        
    \end{tabular}
    \label{tab:fivek}
    \end{center}
\end{table}

\begin{table}[t]
    \centering
    \caption{Quantitative comparison (PPR10K). Models are retrained without extra pretraining data.}
    \begin{tabular}
    {p{20mm}<{\centering} p{14mm}<{\centering} p{14mm}<{\centering} p{14mm}<{\centering} p{14mm}<{\centering} p{14mm}<{\centering} p{14mm}<{\centering}}
        \toprule[1.1pt]
        \multirow{2}{*}{Method} &\multicolumn{2}{c}{PPR10k-a}  & \multicolumn{2}{c}{PPR10k-b} &\multicolumn{2}{c}{PPR10k-c} \\
        \cmidrule(lr){2-3} \cmidrule(lr){4-5} \cmidrule(lr){6-7}
         &PSNR $\uparrow$&$\Delta E \downarrow$&PSNR $\uparrow$&$\Delta E \downarrow$&PSNR $\uparrow$&$\Delta E \downarrow$\\
         \hline
        HDRNet\cite{gharbi2017deep} & 23.93 &8.70 & 23.96 &8.84 & 24.08 &8.87\\
        CSRNet\cite{he2020conditional} &22.72 &9.75&23.76 &8.77& 23.17 &9.45\\
        3D LUT\cite{zeng2020learning} & 24.64 & 8.53& 24.22 &8.33 & 24.10 &7.78\\
        CLUT\cite{zhang2022clut} & 24.89 & 8.33& 24.52 & 8.05&24.51 & 7.55\\
        ICELUT & 24.77 & 8.38& 24.49 & 8.13&24.35 & 7.59\\
        \bottomrule[1.1pt]
    \end{tabular}
    \label{tab:ppk}
\end{table}

\begin{table}[t]
    \centering
    \caption{Runtime (ms) comparison on different hardware platforms. Results tested on 480p images on NVIDIA RTX GeForce 3090, Intel(R) Xeon(R) Platinum 8260L in PC and Cortex-A55 in the low-end smartphone. \textcolor{gray}{Gray} represents the latency or float operations of interpolation. $^\dag$ denotes the original CNN counterpart of ICELUT. - means method unavailable in the platform.}
    \resizebox{0.95\columnwidth}{!}{
    \begin{tabular}
    {p{28mm}<{\centering}  p{20mm}<{\centering}  p{20mm}<{\centering}  p{28mm}<{\centering}  p{18mm}<{\centering}  p{18mm}<{\centering}}
    \toprule[1.1pt]
      \multirow{2}{*}{Method} &\multicolumn{3}{c}{Runtime(ms)}& \multirow{2}{*}{FLOPs(M)} & \multirow{2}{*}{Storage (KB)}  \\
      \cmidrule(lr){2-4}
      &GPU & CPU (PC) & CPU(Smartphone) & \\
      \hline
        UPE\cite{wang2019underexposed} &4.78 &  147.32 & - &143 & 3,996\\
        DPE\cite{chen2018deep} & 23.06 &  327 & - &45,563 & 23,000\\
        HDRNet\cite{gharbi2017deep} & 4.77 &  167.32 & 277 &113 & 1,928\\
        CSRNet\cite{he2020conditional} & 12.14& 357.23 & 504 & 1,268 &148\\
        3DLUT\cite{zeng2020learning} & 1.93+\textcolor{gray}{0.05} & 7.15+\textcolor{gray}{6.71}  & 65.9+\textcolor{gray}{17}&77+\textcolor{gray}{15} & 2,368\\
        CLUT\cite{zhang2022clut} & 2.15+\textcolor{gray}{0.05} & 8.72+\textcolor{gray}{6.71} &141.8+\textcolor{gray}{17}& 75+\textcolor{gray}{15} & 1,168\\
        ICELUT (CNN)$^\dag$ & 2.29 +\textcolor{gray}{0.05} & 10.23+\textcolor{gray}{6.71} & 173.9+\textcolor{gray}{17}&713+\textcolor{gray}{15} & 2,800\\
        ICELUT (LUT) & \textbf{0.35}+\textcolor{gray}{0.05} & \textbf{0.97}+\textcolor{gray}{6.71} & \textbf{7.8}+\textcolor{gray}{17}&\textbf{0.026}+\textcolor{gray}{15} & 780\\
      \bottomrule[1.1pt]
      
    \end{tabular}
    }

    \label{tab:burden}
\end{table}

\subsection{Implementation details}
We use the standard Adam optimizer to minimize the L1 loss function. We set $C=10, L=2, K=5, N=20$ for our backbone and split FC. The mini-batch size is set to 1 and 16 on FiveK and PPR10K, respectively.
To further compress the LUT in color interpolation, we follow the implementation of compressed
representations in \cite{zhang2022clut}. All models are trained for 400 epochs with a fixed learning rate of $1 \times 10^{-4}$.
All experiments are run on an NVIDIA GeForce RTX 3090. 

\subsection{Quantitative results}
We compare our method with the image retouching models: UPE\cite{wang2019underexposed}, DPE\cite{chen2018deep}, HDRNet\cite{gharbi2017deep}, CSRNet\cite{he2020conditional}, 3DLUT\cite{zeng2020learning} and CLUT\cite{zhang2022clut}. The key goal of our work is to achieve competitive results with an extreme inference speed and low energy consumption. Compared to the SOTA method~\cite{zhang2022clut}, ICELUT is on average only 0.2dB lower in FiveK (Table~\ref{tab:fivek}) and PPR10K (Table~\ref{tab:ppk}). Such a slightly lower PSNR can be attributed to the use of all pointwise kernels in ICELUT, which inevitably omits some spatial information to make provision for practical LUT sizes. However, this minor PSNR decrease does not affect the quality of retouched images, which we will explain in the next subsection.

\begin{figure*}[!t]
    \centering
    \includegraphics[width=0.98\textwidth]{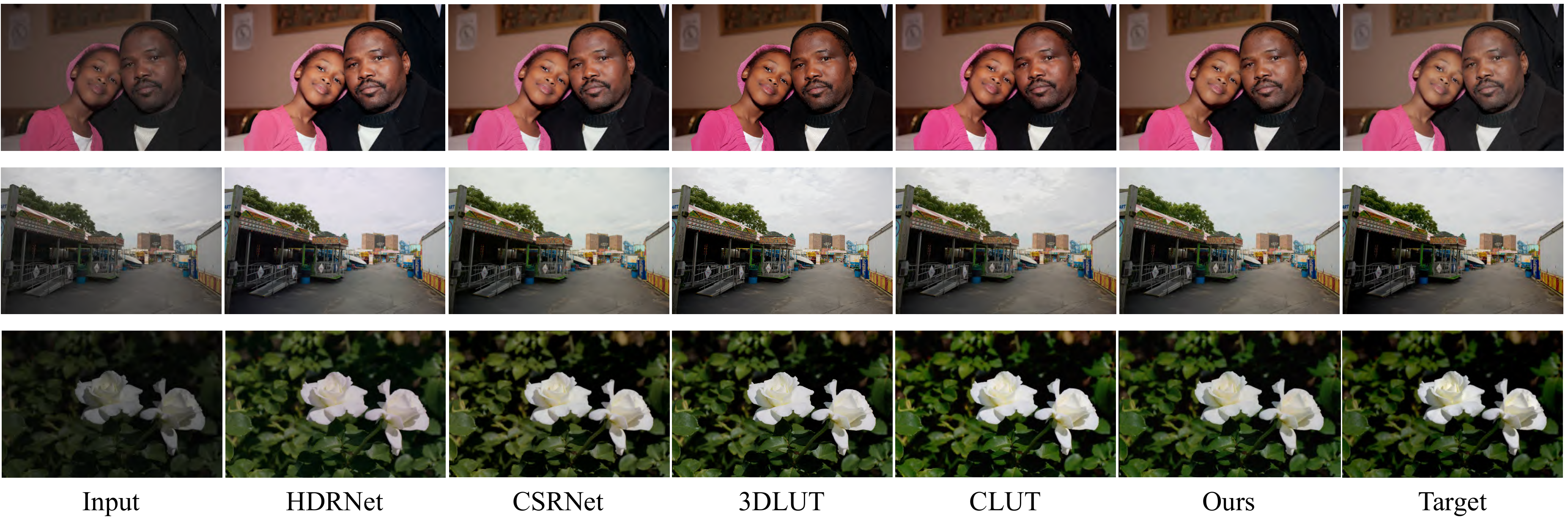}
    \caption{Qualitative comparison of different learning methods for photo retouching on the FiveK.}
    \label{fig:vis}
\end{figure*}


\subsection{Real-time performance comparisons}
To showcase the efficiency of ICELUT for edge devices, we evaluate the inference time and float operations on GPU and CPU. All results are tested on 1,000 480p images and the averaged values are reported. The time measure is conducted on a PC with an Intel(R) Xeon(R) Platinum 8260L, an NVIDIA RTX GeForce 3090 and an entry-level smartphone based on Cortex-A55. As listed in Table~\ref{tab:burden}, ICELUT exceeds all previous methods by a large margin. ICELUT retouches an image in 0.4ms on GPU and 7ms on PC. In particular, we achieve real-time inference on low-end smartphone, which is 6.4$\times$ faster than the current SOTA model. Furthermore, thanks to the efficiency of LUT, ICELUT requires negligible computation to output the weights. We reduce the FLOPs of network from 713M to 26K, which paves the way for saving energy when applied on edge devices. The vast majority of the ICELUT computation comes from the final-step interpolation. Moreover, we remark that ICELUT exhibits 6.1$\times$ fewer FLOPs than 3D LUT~\cite{zeng2020learning} and CLUT~\cite{zhang2022clut}. More importantly, the network's convolution operations have been replaced by extremely cost-effective table lookup operations, reducing the original FLOPs of over 70 million to just 26K. Finally, the storage requirement for ICELUT is less than 1MB, making it memory-friendly for edge devices.


\begin{table}[t]
    \centering
    \caption{Effects of hyper-parameter $K$ on the enhancement performance. Note that $\dag$ means only the pooled features from the last layer of the backbone are used for weight prediction. Size denotes the storage of Weight LUTs.}
    \begin{tabular}
    {p{16mm}<{\centering}  p{6mm}<{\centering}  p{6mm}<{\centering}  p{6mm}<{\centering}  p{18mm}<{\centering}  p{16mm}<{\centering}  p{16mm}<{\centering}}
        \toprule[1.1pt]
        Method & C & K & L &PSNR$\uparrow$ & SSIM$\uparrow$ &Size (KB)\\
        \hline

         \multirow{2}{*}{SFC} &6&3 & 2&24.88 & 0.897 & 240 \\
         &6&2 & 3&25.00 & 0.904 & 10,240 \\
         FC &6&- & - & 25.03 & 0.903  & - \\
                 \hline
         \multirow{2}{*}{SFC} &12&6 & 2&25.27 & 0.911 & 480\\
         &12&4 & 3&25.28 & 0.912 & 20,480\\
         FC &12&- & - & 25.31 & 0.914  & - \\
                 \hline
         \multirow{2}{*}{SFC}&18&9 & 2&25.33 & 0.916& 1,440\\
         &18&6 & 3&25.36 & 0.918 & 30,720\\
         FC &18&- & - & 25.38 & 0.919 & - \\
         \hline
         \multirow{3}{*}{Pooling$^\dag$} &6&- & -&21.98 & 0.766 & - \\
          &12&- & -&22.11 & 0.769 & - \\
          &18&- & -&22.29 & 0.774 & - \\
         
        \bottomrule[1.1pt]
        
    \end{tabular}
    \label{tab:sfc}
\end{table}

\subsection{Ablation study}
We conduct ablation studies on the FiveK dataset (480p) to verify the critical components in ICELUT.

\subsubsection{Split FC layer}\
As described earlier, a split FC layer is employed to save memory while achieving global information fusion effectively. Here, we give the ablation study of this layer. We set the FC output dimension to 20, i.e., 20 weights for weighting 20 basis LUTs. First, we quantize the output from FP32 to INT8 for saving the LUT memory. Note that the quantization error has little impact on the performance. To quantitatively demonstrate the impact on the overall performance and determine the most suitable settings, we first evaluate with the channel numbers $C=\{6, 12, 18\}$ and split group lengths $L=\{2,3\}$ (for the reason that $L>3$ already leads to an undesirable LUT size). We remark $C = K\times L$, where $K$ is the SFC number. The results are in Table~\ref{tab:sfc}. It can be seen that an increase in $K$ leads to improved model performance. To strike a balance between memory size and performance, we choose $C=18$ for subsequent experiments. Compared to a vanilla FC layer, which gathers full channel features for predictions, it is evident that substituting the SFC layer with an FC layer yields only marginal improvements at the expense of a much higher storage cost. Furthermore, we compare our method with the model using pooled features for weight prediction in the last row of Table~\ref{tab:sfc}. Obviously, without the global fusion, such as the FC or SFC layers, the features extracted from the pointwise convolution network are insufficient to achieve high performance. 

\begin{table}[!h]
    \centering
    \caption{Quantization strategy. Size denotes the storage of the LUTs.}
    \begin{tabular}{p{5mm}<{\centering}  p{5mm}<{\centering}  p{8mm}<{\centering}  p{8mm}<{\centering}  p{8mm}<{\centering}  p{30mm}<{\centering} p{22 mm}<{\centering} p{18 mm}<{\centering}}
    \toprule[1.1pt]
        \# & $\Delta s$& R & I & C & Weight LUT Size (KB)& Total Size (KB) &PSNR/dB\\
        \hline
        0 & 4 & 32 & 256 & 20& 12,800 &13,500 & 25.35\\
        1 & 4 & 32 & 256 & 10& 6,400 &6,780 & 25.32\\
        2 & 2 & 16 & 64 & 20 & 800 & 1,500& 25.30\\
        3 & 2 & 16 & 64 & 10 & 400 & 780&25.27 \\
        4 & 2 & 8 & 32 & 20 & 200 & 900&24.77 \\
        \bottomrule[1.1pt]
    \end{tabular}
    \label{tab:wlut}
\end{table}

\subsubsection{Quantization of Weight LUTs for downscaling storage}
It is crucial for application in edge devices with comparable small storage size. The storage of Weight LUTs is a majority component of the whole size. In Sec.~\ref{sec:sfc}\&~\ref{sec:sfc2}, we have discussed the factors of Weight LUTs. Here we explore the quantization strategy for Weight LUTs and the influence on storage. In Table~\ref{tab:wlut}, $I$ represents the potential index values in Weight LUTs. Increasing $I$ results in a smaller sampling interval, thereby reducing quantization error. The parameter $C$ corresponds to the output channel of the pooling module. As $C$ increases, the feature becomes richer, leading to a more detailed color information. For our model, we have chosen $I=64$ and $C=10$.


\section{Conclusion}
This work has proposed the first-of-its-kind purely LUT-based image enhancer, ICELUT, for extremely low-cost and high-speed image retouching. We reveal that input RGB channels are vital for performance and employ fully pointwise convolution kernels that favor subsequent LUT conversion after training. A novel split fully connected layer is devised to effectively suppress the LUT size without compromising performance. 
With the concerted efforts of these novel designs, our purely LUT-based scheme achieves a remarkable 0.4ms (7ms) on GPU (CPU) with near-state-of-the-art performance, and reduces the power consumption to a negligible level compared to all other CNN solutions.

\section*{Acknowledgements}
This work was partly supported by the National Natural Science Foundation of China (Grant No. 61991451), the Shenzhen Science and Technology Program (JCYJ20220818101001004), the Theme-Based Research
Scheme (TRS) Project T45-701/22-R and General Research Fund (GRF) Project 17203224 of the Research Grants Council (RGC), HKSAR.

%
%
\bibliographystyle{splncs04}
\bibliography{main}

\begin{thebibliography}{10}
\providecommand{\url}[1]{\texttt{#1}}
\providecommand{\urlprefix}{URL }
\providecommand{\doi}[1]{https://doi.org/#1}

\bibitem{afifi2019color}
Afifi, M., Price, B., Cohen, S., Brown, M.S.: When color constancy goes wrong: Correcting improperly white-balanced images. In: Proceedings of the IEEE/CVF conference on computer vision and pattern recognition. pp. 1535--1544 (2019)

\bibitem{backhaus2011color}
Backhaus, W.G., Kliegl, R., Werner, J.S.: Color vision: Perspectives from different disciplines. Walter de Gruyter (2011)

\bibitem{bychkovsky2011learning}
Bychkovsky, V., Paris, S., Chan, E., Durand, F.: Learning photographic global tonal adjustment with a database of input/output image pairs. In: CVPR 2011. pp. 97--104. IEEE (2011)

\bibitem{chen2018deep}
Chen, Y.S., Wang, Y.C., Kao, M.H., Chuang, Y.Y.: Deep photo enhancer: Unpaired learning for image enhancement from photographs with gans. In: Proceedings of the IEEE conference on computer vision and pattern recognition. pp. 6306--6314 (2018)

\bibitem{gharbi2017deep}
Gharbi, M., Chen, J., Barron, J.T., Hasinoff, S.W., Durand, F.: Deep bilateral learning for real-time image enhancement. ACM Transactions on Graphics (TOG)  \textbf{36}(4),  1--12 (2017)

\bibitem{he2020conditional}
He, J., Liu, Y., Qiao, Y., Dong, C.: Conditional sequential modulation for efficient global image retouching. In: Computer Vision--ECCV 2020: 16th European Conference, Glasgow, UK, August 23--28, 2020, Proceedings, Part XIII 16. pp. 679--695. Springer (2020)

\bibitem{he2016deep}
He, K., Zhang, X., Ren, S., Sun, J.: Deep residual learning for image recognition. In: Proceedings of the IEEE conference on computer vision and pattern recognition. pp. 770--778 (2016)

\bibitem{huang2023hundred}
Huang, B., Li, J.C.L., Ran, J., Li, B., Zhou, J., Yu, D., Wong, N.: Hundred-kilobyte lookup tables for efficient single-image super-resolution. arXiv preprint arXiv:2312.06101  (2023)

\bibitem{jo2021practical}
Jo, Y., Kim, S.J.: Practical single-image super-resolution using look-up table. In: Proceedings of the IEEE/CVF Conference on Computer Vision and Pattern Recognition. pp. 691--700 (2021)

\bibitem{kim2020global}
Kim, H.U., Koh, Y.J., Kim, C.S.: Global and local enhancement networks for paired and unpaired image enhancement. In: Computer Vision--ECCV 2020: 16th European Conference, Glasgow, UK, August 23--28, 2020, Proceedings, Part XXV 16. pp. 339--354. Springer (2020)

\bibitem{kim2020pienet}
Kim, H.U., Koh, Y.J., Kim, C.S.: Pienet: Personalized image enhancement network. In: Computer Vision--ECCV 2020: 16th European Conference, Glasgow, UK, August 23--28, 2020, Proceedings, Part XXX 16. pp. 374--390. Springer (2020)

\bibitem{li2022mulut}
Li, J., Chen, C., Cheng, Z., Xiong, Z.: Mulut: Cooperating multiple look-up tables for efficient image super-resolution. In: European Conference on Computer Vision. pp. 238--256. Springer (2022)

\bibitem{liang2021ppr10k}
Liang, J., Zeng, H., Cui, M., Xie, X., Zhang, L.: Ppr10k: A large-scale portrait photo retouching dataset with human-region mask and group-level consistency. In: Proceedings of the IEEE/CVF Conference on Computer Vision and Pattern Recognition. pp. 653--661 (2021)

\bibitem{liu20234d}
Liu, C., Yang, H., Fu, J., Qian, X.: 4d lut: learnable context-aware 4d lookup table for image enhancement. IEEE Transactions on Image Processing  \textbf{32},  4742--4756 (2023)

\bibitem{ma2022learning}
Ma, C., Zhang, J., Zhou, J., Lu, J.: Learning series-parallel lookup tables for efficient image super-resolution. In: European Conference on Computer Vision. pp. 305--321. Springer (2022)

\bibitem{moran2021curl}
Moran, S., McDonagh, S., Slabaugh, G.: Curl: Neural curve layers for global image enhancement. In: 2020 25th International Conference on Pattern Recognition (ICPR). pp. 9796--9803. IEEE (2021)

\bibitem{ouyang2023rsfnet}
Ouyang, W., Dong, Y., Kang, X., Ren, P., Xu, X., Xie, X.: Rsfnet: A white-box image retouching approach using region-specific color filters. In: Proceedings of the IEEE/CVF International Conference on Computer Vision. pp. 12160--12169 (2023)

\bibitem{shih2013data}
Shih, Y., Paris, S., Durand, F., Freeman, W.T.: Data-driven hallucination of different times of day from a single outdoor photo. ACM Transactions on Graphics (TOG)  \textbf{32}(6),  1--11 (2013)

\bibitem{wang2019underexposed}
Wang, R., Zhang, Q., Fu, C.W., Shen, X., Zheng, W.S., Jia, J.: Underexposed photo enhancement using deep illumination estimation. In: Proceedings of the IEEE/CVF conference on computer vision and pattern recognition. pp. 6849--6857 (2019)

\bibitem{wang2021real}
Wang, T., Li, Y., Peng, J., Ma, Y., Wang, X., Song, F., Yan, Y.: Real-time image enhancer via learnable spatial-aware 3d lookup tables. In: Proceedings of the IEEE/CVF International Conference on Computer Vision. pp. 2471--2480 (2021)

\bibitem{wang2004image}
Wang, Z., Bovik, A.C., Sheikh, H.R., Simoncelli, E.P.: Image quality assessment: from error visibility to structural similarity. IEEE transactions on image processing  \textbf{13}(4),  600--612 (2004)

\bibitem{yang2022seplut}
Yang, C., Jin, M., Xu, Y., Zhang, R., Chen, Y., Liu, H.: Seplut: Separable image-adaptive lookup tables for real-time image enhancement. In: European Conference on Computer Vision. pp. 201--217. Springer (2022)

\bibitem{zeng2020learning}
Zeng, H., Cai, J., Li, L., Cao, Z., Zhang, L.: Learning image-adaptive 3d lookup tables for high performance photo enhancement in real-time. IEEE Transactions on Pattern Analysis and Machine Intelligence  \textbf{44}(4),  2058--2073 (2020)

\bibitem{zhang2022clut}
Zhang, F., Zeng, H., Zhang, T., Zhang, L.: Clut-net: Learning adaptively compressed representations of 3dluts for lightweight image enhancement. In: Proceedings of the 30th ACM International Conference on Multimedia. pp. 6493--6501 (2022)

\end{thebibliography}

\title{Taming Lookup Tables for Efficient Image Retouching Supplementary Material} 
\author{}
\institute{}
\maketitle

\section{Low Input Resolution Inference}
\begin{figure}[htp]
    \centering
    \includegraphics[width=0.98\textwidth]{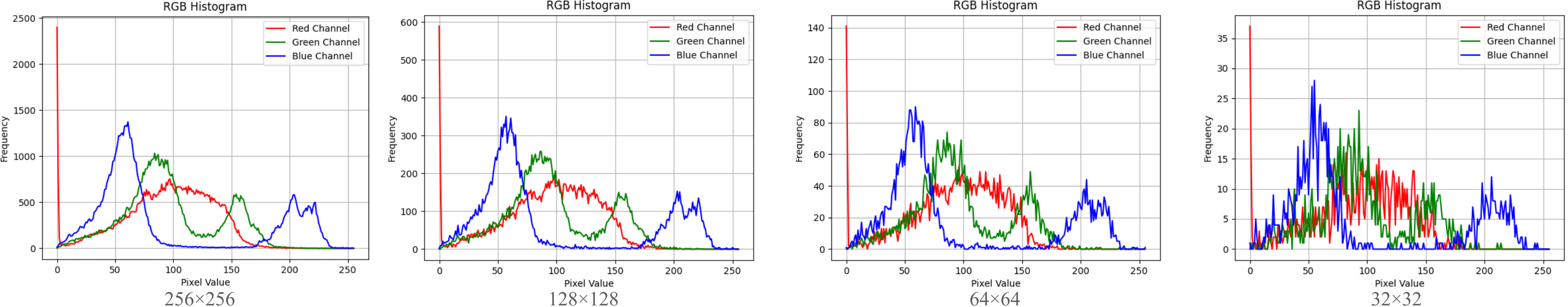}
    \caption{Histogram of images at different resolutions.}
    \label{fig:hist}
\end{figure}

The reasons for these results that networks with a larger RF suffer from a more severe
performance drop in Sec.~3.4 can be attributed to two main factors. First, both 3DLUT and CLUT are composed of multiple $3\times3$ CNN layers, thus inheriting the inductive bias inherent in the large kernels. This results in a significant performance drop when the scales of downsampled images are inconsistent with the training images. Second, all kernels in our method are $1\times1$, so the interdependence among spatial pixels is decoupled. This enables our method to achieve robust performance at different scales. While it is apparent that downsized images may lose information, when visualizing histograms of different image scales in Fig.~\ref{fig:hist}, we observe that low-resolution images still retain the fundamental color information of the original high-resolution images. This explains that the network can still capture vital image attributes, including brightness, color, and tonal characteristics, and make correct predictions for LUT weights.

\section{Experiments}
\subsection{Evaluation metrics}
We employ three metrics, including peak signal-to-noise ratio (PSNR), structural similarity index (SSIM)~\cite{wang2004image}, and $\Delta E$ to evaluate different methods. $\Delta E$ is a color difference metric defined in the CIELAB color space~\cite{backhaus2011color}.

\subsection{Failure case}
\begin{figure}[ht!] 
    \centering
    \includegraphics[width=0.98\textwidth]{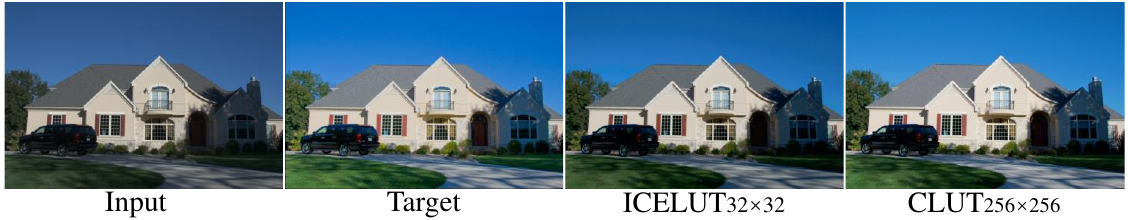}
     \caption{Failure case.}
     \label{fig2}
\end{figure}

Our method generates unsatisfying results on images with large high-contrast smoothing regions, as shown in the Fig.~\ref{fig2}. Because average pooling compresses the spatial size into $1 \times 1$, the large smooth regions will suffer from global biased tone or brightness from other regions due to the average operation. Nonetheless, such cases occupy only 1.5\% of the whole dataset, and are still visually plausible as shown in Fig.~\ref{fig2}.

\vspace{-2em}

\section{Ablation Study}
\subsection{Split FC}The interactions between channels have largely been captured by pointwise convolution (whose wide channel dimension during training will not burden inference), diminishing the significance of FC in establishing global channel connections. It means we could use a broader convolution network as our backbone with the sack of no inference burden. Hence, a weakened version, split FC, is already capable of transforming image information into basis coefficients with a powerful backbone, as shown in Table~\ref{tab:sfc_burden} whose middle layer has the largest dimension.

\begin{table}[htp]
    \centering
    \caption{Comparison between different backbones with vanilla FC or split FC.}
    \begin{tabular}
    {p{50mm}<{\centering}  p{30mm}<{\centering}  p{30mm}<{\centering}} 
        \toprule[1.1pt]
       Backbone Dimension & PSNR (w/ FC) & PSNR (w/ split FC)\\
        \hline
        32-64-128-256-512-256-128-64-32 &  25.33 & 25.31 \\
        16-32-64-128-256-128-64-32-16 & 25.29 & 25.14 \\
        \bottomrule[1.1pt]
    \end{tabular}
    \label{tab:sfc_burden}
\end{table}

\subsection{High-low bit separation}
To efficiently construct the Channel LUT for storing the output of each input pixel, we employ two parallel branches to process the MSBs and LSBs independently, resulting in significant LUT memory savings compared to the whole INT8 input. Again, we set the output channel number $C=10$. As depicted in Table~\ref{tab:two}, while there is a very slight drop in performance, the parallel design substantially reduces memory consumption by a factor of $1500\times$ vs a single-branch architecture.

\begin{table}[htp]
    \centering
    \caption{Comparison of one-branch (8-bit input) and two-branch (two 4-bit inputs) networks.}
    \begin{tabular}
        {p{24mm}<{\centering}  p{16mm}<{\centering}  p{16mm}<{\centering}  p{32mm}<{\centering}}
        \toprule[1.1pt]
       Method&PSNR  & SSIM & Channel LUT Size (MB) \\
            \hline
        Single (8 bits) & 25.35 & 0.919& 1,342\\
        Parallel (4 bits) & 25.27 &0.918 & 0.20\\
      \bottomrule[1.1pt]
    \end{tabular}
    \label{tab:two}
\end{table}

\subsection{Number of basis LUTs}
To investigate the effect of the number of basis LUTs $N$, we set $C=10, K=5, L=2$ and $N=\{1, 3, 5,10,15,20\}$, and compute the sizes of LUTs converted from the SFC layer in Table~\ref{tab:my_label}. Notably, when $N$ is small, the expressiveness of color transforms is restricted by the limited representation of basis LUTs. When $N$ is large enough, the performance improvement is only minor.

\begin{table}[htp]
    \centering
        \caption{Effect of the number $N$ of basis LUTs. Size denotes the storage of Weight LUTs.}
    \begin{tabular}{p{18mm}<{\centering}  p{10mm}<{\centering}  p{10mm}<{\centering}  p{10mm}<{\centering}  p{10mm}<{\centering}  p{10mm}<{\centering} p{10 mm}<{\centering}}
    \toprule[1.1pt]
    $N$ & 1 & 5 & 10 & 15 & 20 & 25\\
    \hline
    PSNR (dB) & 23.27 & 25.16 & 25.27 & 25.29 & 25.30 &25.28\\ 
    Size (KB) & 40 & 200 & 400 & 600 & 800 & 1,000\\ 
        \bottomrule[1.1pt]
    \end{tabular}
    \label{tab:my_label}
\end{table}

\end{document}